\documentclass{article}

\usepackage{microtype}
\usepackage{graphicx}
\usepackage{subcaption}
\usepackage{booktabs} 

\usepackage{hyperref}
\usepackage{enumitem}
\usepackage{multirow}

\usepackage[accepted]{icml2026}

\usepackage{amsmath}
\usepackage{amssymb}
\usepackage{mathtools}
\usepackage{amsthm}

\usepackage[table]{xcolor}

\definecolor{sinkblue}{HTML}{EAF2FB}
\definecolor{nullgreen}{HTML}{EEF7EE}
\definecolor{driftorange}{HTML}{FFF3E0}
\definecolor{controlgray}{HTML}{F5F5F5}
\definecolor{softyellow}{HTML}{FFF8E1}

\newcommand{\sinkcell}[1]{\cellcolor{sinkblue}\textbf{#1}}
\newcommand{\nullcell}[1]{\cellcolor{nullgreen}#1}
\newcommand{\driftcell}[1]{\cellcolor{driftorange}\textbf{#1}}
\newcommand{\ctrlcell}[1]{\cellcolor{controlgray}#1}

\usepackage[capitalize,noabbrev]{cleveref}

\theoremstyle{plain}

\theoremstyle{definition}

\theoremstyle{remark}

\usepackage[textsize=tiny]{todonotes}

\icmltitlerunning{Attention Sinks in Diffusion Transformers: A Causal Analysis}

\begin{document}

\twocolumn[
  \icmltitle{Attention Sinks in Diffusion Transformers: \\A Causal Analysis}



  \icmlsetsymbol{equal}{*}

    \begin{icmlauthorlist}
    \icmlauthor{Fangzheng Wu}{tulane}
    \icmlauthor{Brian Summa}{tulane}
    \end{icmlauthorlist}
    
    \icmlaffiliation{tulane}{Department of Computer Science, Tulane University, New Orleans, LA, USA}
    
    \icmlcorrespondingauthor{Fangzheng Wu}{fwu6@tulane.edu, fwu66666666@gmail.com}

  \icmlkeywords{Machine Learning, ICML}

  \vskip 0.3in
]



\printAffiliationsAndNotice{}  

\begin{figure*}[t]
  \centering
  \includegraphics[width=0.9\textwidth]{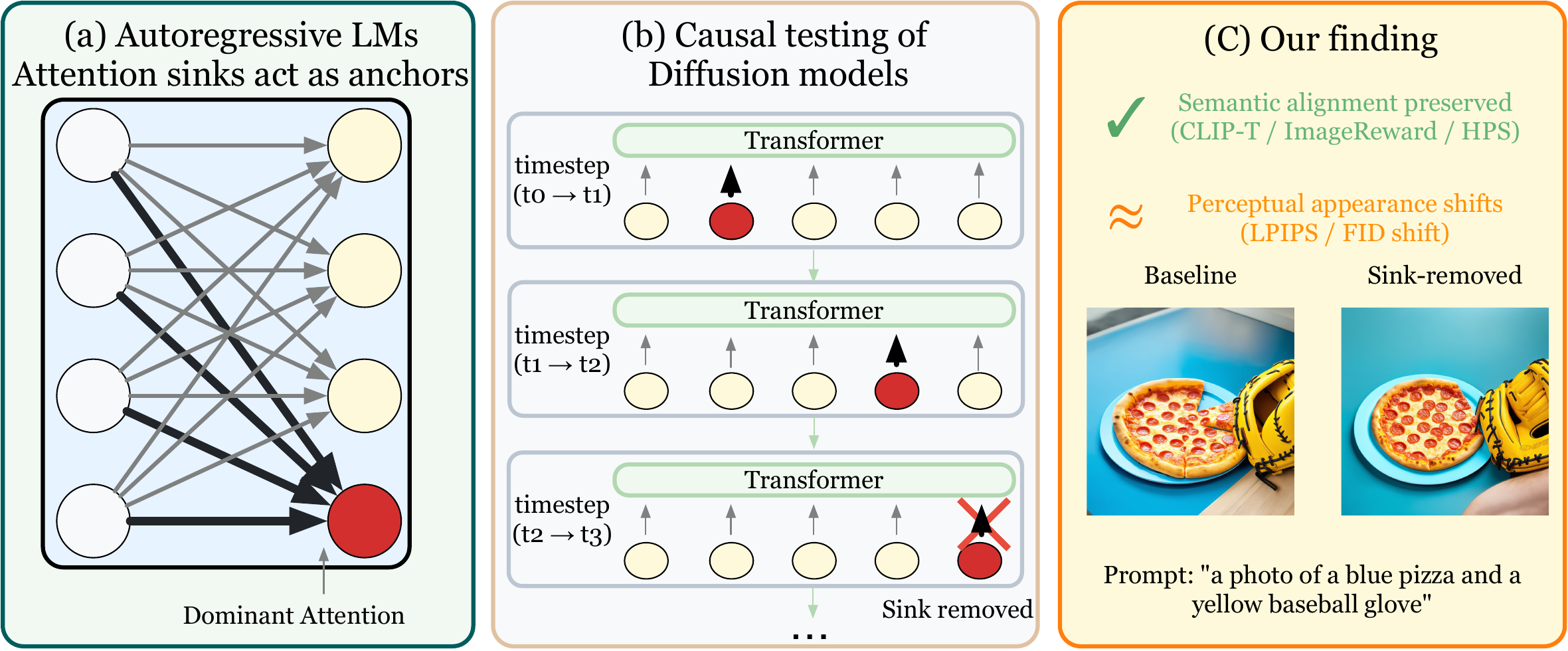}
  \caption{\textbf{Attention sinks in diffusion transformers.}
  \textbf{(a)} In autoregressive LMs, attention sinks often act as stable anchors that attract dominant attention mass.
  \textbf{(b)} In diffusion transformers, dominant recipients vary across denoising timesteps; we perform a \emph{causal} test by dynamically identifying sink tokens per step and suppressing them during inference.
  \textbf{(c)} Sink suppression preserves semantic alignment and preference scores (CLIP-T / ImageReward / HPS-v2), yet can induce perceptual and distributional shifts relative to baseline outputs (LPIPS / FID$_\text{shift}$), consistent with moving samples within the model's output manifold.}
  \label{fig:teaser}
\end{figure*}

\begin{abstract}
Attention sinks---tokens that receive disproportionate attention mass---are 
assumed to be functionally important in autoregressive language models, but 
their role in diffusion transformers remains unclear.
We present a causal analysis in text-to-image diffusion, dynamically identifying 
dominant attention recipients per timestep and suppressing them via paired, 
training-free interventions on the score and value paths.
Across 553 GenEval prompts on Stable Diffusion~3 (with SDXL corroboration), 
removing these sinks does not degrade text-image alignment (CLIP-T) or 
preference proxies (ImageReward, HPS-v2) at $k{=}1$; only under stronger 
interventions ($k\!\geq\!10$) does HPS-v2 exhibit a metric-dependent boundary, 
while CLIP-T remains robust throughout.
The perceptual shifts induced by suppression are nonetheless 
\emph{sink-specific}---$\sim\!6\times$ larger than equal-budget random 
masking---revealing an empirical dissociation between trajectory-level 
perturbation and \emph{semantic alignment} in diffusion transformers.
\footnote{Code 
available at \url{https://github.com/wfz666/ICML26-attention-sink}.}
\end{abstract}

\section{Introduction}

Attention mechanisms in large language models frequently converge on specific \emph{attention sinks}—tokens or positions that accumulate disproportionate attention mass independently of their semantic importance~\citep{xiao2024streamingllm,sun2024massiveactivations}.
In autoregressive language models, such sinks have been interpreted as
functionally important structures: they serve as stable anchors for
key-value caching in long-context generation, prevent attention entropy
collapse, and act as implicit registers for residual information.
Empirically, removing or disrupting sinks in autoregressive models can
degrade generation quality or destabilize inference.
As a result, attention sinks are often treated not as statistical 
artifacts but as load-bearing computational structures in 
transformer-based generation.

However, it is unclear whether this intuition transfers to diffusion
transformers.
Unlike autoregressive decoding, which proceeds token-by-token under
causal masking, diffusion models perform non-causal, bidirectional
attention across multiple denoising steps, iteratively refining all
positions simultaneously rather than committing to irreversible
left-to-right decisions~\citep{peebles2022dit,saharia2022photorealistic}.
In this setting, attention does not obviously serve as a persistent memory anchor for preceding context (a role often attributed to sinks in autoregressive settings).
This architectural distinction raises a natural question:
\emph{are attention sinks equally necessary under diffusion-style
inference, or does the intuition from autoregressive models fail to transfer in practice?}
The answer matters concretely for efficient diffusion attention design, 
especially in DiT-style architectures. Recent sparsification and 
attention-compression methods make different implicit assumptions about 
high-mass attention recipients: some preserve tokens through 
attention-driven importance scores~\citep{wang2024atedm}, whereas others 
impose locality or structured sparsity patterns~\citep{yuan2024ditfastattn,ren2025grat,sparsedit2025} 
that may discard such recipients depending on the layer and timestep. 
This leaves open a basic but consequential ambiguity: does high incoming 
attention mass indicate functional necessity, or merely mark a replaceable 
routing pattern? Without causal evidence on sink necessity, these design 
choices lack grounding for which high-mass recipients to preserve.

Existing reports of sink-like phenomena in diffusion-style models~\citep{arriola2025block,wen2025analysis,rulli2025diffusionlm} 
suggest some dominant recipients are removable without catastrophic failure, but the 
evidence is fragmented---definitions vary across studies (fixed-position vs.\ dynamic), 
interventions are often observational rather than causal, and evaluations are limited 
to specific modalities or small-scale settings. Whether attention sinks are truly 
\emph{necessary} for high-quality generation in text-to-image diffusion transformers 
thus remains unresolved.

In this work, we provide a systematic causal analysis of attention sinks
in diffusion transformers (DiT)~\cite{peebles2022dit}.
We define sinks dynamically as key positions receiving the largest incoming attention mass, separately for each attention head and denoising timestep. This approach moves beyond the fixed-position assumptions typical of autoregressive settings, which our empirical data suggest are largely invalid (index-0 overlap $<$0.2\%).
Using paired, training-free interventions along both the score (logit)
and value paths, we test the necessity of sinks across layers, denoising
phases, intervention intensities, and architectures (SD3~\cite{esser2024scaling} and SDXL~\cite{sdxl}).
All experiments employ strict seed-matched generation with bootstrap
confidence intervals, ensuring that observed differences reflect
intervention effects rather than sampling variance.

Across large-scale evaluations on 553 prompts from GenEval~\cite{geneval} with
Stable Diffusion~3, and corroborating experiments on 5,000 COCO captions
and SDXL, we find that removing dynamically identified attention sinks
does not degrade semantic alignment (CLIP-T~\citep{hessel2021clipscore}) or preference metrics
(ImageReward~\cite{imagereward}, HPS-v2~\cite{hpsv2}), which are trained on large-scale human 
preference datasets such as Pick-a-Pic~\citep{kirstain2023pickapic}.
At the same time, suppressing dominant recipients induces perceptual shifts 
that are \emph{sink-specific}---roughly $6\times$ larger than equal-budget 
random masking---revealing an empirical dissociation between trajectory-level 
perturbation and alignment-level robustness (Figure~\ref{fig:teaser}).
Together, these findings clarify that attention sinks are not functionally
necessary for semantic alignment in diffusion transformers, while
identifying sink-specific perceptual drift as a structural boundary 
condition of sink suppression.

More broadly, our results provide causal evidence that incoming attention 
mass---widely used as a proxy for token importance---does not reliably 
predict functional necessity for semantic alignment in diffusion transformers.
We focus on alignment necessity at the level of widely used proxy metrics 
(CLIP-T, ImageReward, HPS-v2), treating perceptual changes as boundary 
conditions and additionally characterizing a metric-dependent boundary 
under stronger interventions where preference proxies show sink-specific 
effects while alignment remains stable. Skill-based and holistic evaluation 
frameworks---such as Gecko~\citep{gecko2024} and HEIM~\citep{lee2023heim}---that 
probe finer compositional fidelity remain a natural complement for future work.

\paragraph{Conflict of Interest Disclosure.}
The authors have no financial or other substantive conflicts of interest to disclose.

\section{Related Work}

\textbf{Attention sinks in autoregressive language models.}
In autoregressive (AR) language models, certain tokens---often the 
beginning-of-sequence token or early positions---receive 
disproportionately high attention mass regardless of their semantic 
relevance~\citep{xiao2024streamingllm,sun2024massiveactivations}.
These ``attention sinks'' have been linked to several functional roles: 
serving as stable anchors for key-value caching in long-context 
generation~\citep{xiao2024streamingllm}, preventing attention entropy 
collapse~\citep{sun2024massiveactivations}, and acting as implicit 
registers that accumulate residual information.
Similar ``register'' phenomena have been observed in discriminative
vision transformers~\citep{darcet2024registers}, raising the question
of whether such structures serve analogous functions in generative
diffusion transformers. \citet{jiang2025registers} subsequently show that the high-norm
activations responsible for register-like behavior can be relocated
at inference into an additional untrained token, achieving the effect
of trained registers without retraining; this suggests register-style
structures may be less load-bearing than initially thought even in
discriminative settings, complementing our finding of sink
dispensability in the generative diffusion regime.
Empirically, removing or disrupting sinks in AR models can degrade
generation quality or destabilize long-context 
performance~\citep{gu2024attention}.
These findings shaped a widespread intuition that attention sinks are 
functionally necessary structures.
\textbf{A common but largely untested assumption is that this 
functional role transfers to diffusion-style bidirectional denoising.}

\textbf{Sink-like phenomena in diffusion-based models.}
Recent work has begun to probe attention concentration in 
non-autoregressive settings, with results that diverge from the AR 
narrative.
In discrete diffusion language models, \citet{arriola2025block}
observe sink-like attention patterns and report that certain dominant
recipients can be ablated without catastrophic degradation, suggesting
a potential commonality across discrete and continuous diffusion regimes,
though systematic causal verification remains limited.
Concurrent work by \citet{rulli2025diffusionlm} reports strikingly
parallel findings in masked diffusion language models: sinks exhibit
dynamic positions throughout generation and remain largely robust to
suppression, providing independent convergent evidence in the
discrete-token regime that sink-like phenomena recur across diffusion
architectures.
For video diffusion transformers, \citet{wen2025analysis} identify
redundant attention heads and demonstrate that certain connections 
can be pruned, with caveats regarding layer sensitivity.
Critically, definitions of sinks vary across these studies: 
AR models exhibit \emph{positionally anchored} sinks tied to 
structural tokens (e.g., BOS), whereas diffusion models may exhibit 
\emph{dynamically varying} concentration that shifts across timesteps.
Our finding that index-0 overlap is negligible ($<$0.2\%, often $<$0.1\% 
in our measurements) confirms this distinction and underscores the need 
for dynamic sink definitions in diffusion settings.

\textbf{Attention sparsification and acceleration.}
Recent work explores attention compression for diffusion transformers, 
including window attention~\citep{yuan2024ditfastattn}, token 
pruning~\citep{wang2024atedm}, linear attention~\citep{xie2024sana}, 
and phase-aware caching~\citep{zhao2024pab,liu2024faster}.
These methods demonstrate that sparsity can be beneficial, 
\emph{yet they do not isolate whether dominant recipients are 
causally necessary}.
Our work provides the missing causal evidence, complementing recent
structured sparsification approaches such as GRAT~\citep{ren2025grat}.
Linear- and gated-linear-attention diffusion variants~\citep{meng2025norm,zhu2025dig}
modify the attention operator itself rather than the recipients within
softmax attention. Our conclusions are scoped to standard
softmax-attention diffusion architectures; we do not claim the same
necessity profile holds under alternative attention mechanisms
(e.g., linear attention).


\textbf{Causal diagnostics at the head level.}
Complementary to token-level analysis, prior work has studied 
which \emph{attention heads} matter for generation.
\citet{michel2019sixteen} show that many heads can be pruned 
with minimal quality loss in language models.
In diffusion settings, head-level gating has been explored for 
efficiency~\citep{li2024snapfusion}.
These approaches ask ``which heads matter''; we pursue an 
\emph{orthogonal} inquiry: ``do the most attended tokens within 
active heads matter?''
Our results suggest that dominant attention recipients are 
non-functional even in heads that remain active, suggesting that 
head-level and token-level redundancy may be complementary and 
warrant separate investigation.

\textbf{Attention editing and modulation in diffusion models.}
A separate line of work manipulates attention patterns to improve 
text--image alignment and compositional fidelity~\citep{tang2022daam}.
Attend-and-Excite~\citep{chefer2023attendandexcite} amplifies 
cross-attention to neglected tokens during inference, improving 
object presence.
Earlier attention editing frameworks such as Prompt-to-Prompt~\citep{hertz2022prompt2prompt}
demonstrate that selectively manipulating cross-attention can steer semantic content 
without retraining.
More recent attention regulation methods explicitly reweight dominant attention 
patterns to improve compositional fidelity~\citep{zhang2024attentionreg}.
Structured cross-attention methods~\citep{feng2023trainingfree} 
decompose prompts to reduce attribute binding errors.
More recently, phase-aware attention modulation has been explored 
for layout control~\citep{chen2024training}.
These methods demonstrate that \emph{modifying} dominant attention 
can improve alignment; our work addresses the complementary question 
of whether such dominant recipients are \emph{necessary} for alignment.
Our results suggest they are not: suppressing high-mass recipients 
preserves alignment metrics under standard settings, indicating a clear 
distinction between \emph{usefulness for optimization or controllability} 
and \emph{necessity for semantic fidelity}.

\section{Experiments and Results}

\subsection{Experimental Setup}

\paragraph{Models and evaluation.}
We use SD3 (joint attention over image/text tokens) as our primary testbed 
and validate on SDXL (U-Net cross-attention); all models are used in 
inference mode without finetuning.
We evaluate on 553 GenEval prompts using strict seed-paired generation, 
reporting mean paired differences with 95\% bootstrap CIs (1k resamples).
Supplementary analyses use smaller prompt subsets; details are in 
Appendix~\ref{app:appendix_setup}.

\paragraph{Metrics.}
We use CLIP-T~\citep{hessel2021clipscore} as the primary alignment metric, 
with ImageReward and HPS-v2 as preference proxies.
Perceptual shifts are quantified via LPIPS and FID$_{\text{shift}}$ 
(distributional distance between baseline and intervention outputs, rather than against real images, as a sample-domain measure of distributional drift caused by the 
intervention.).
We adopt $|\Delta|<0.002$ for CLIP-T as the practical equivalence margin, 
chosen relative to the empirical noise floor observed under seed variation 
and bootstrap uncertainty at $N{=}553$; further justification is in Appendix~\ref{app:appendix_setup}.

\paragraph{Interventions.}
We suppress sinks via two pathways: (1)~\textbf{score-path}: add 
$\log\eta$ to sink logits ($\eta{=}0$ effectively zeros attention); 
(2)~\textbf{value-path}: replace sink value vectors with alternatives 
(zero, mean, or interpolation).
Both are training-free and applied during inference.
Consistent conclusions across both pathways suggest robustness beyond
any single masking scheme.

We emphasize that our claims concern \emph{inference-time aggregation
necessity}---whether suppressing sink tokens during the forward pass
degrades the evaluated metrics---and are distinct from training-time
functional roles such as gradient stabilization, which we do not test.

\subsection{H1: Dynamics of Attention Sinks}
\label{sec:h1}

We first characterize the \emph{existence}, \emph{layer-wise distribution}, and \emph{temporal dynamics} of attention sinks during diffusion sampling.

\paragraph{Dynamic Sink Definition.}
Rather than assuming a fixed sink position (e.g., index-0 as in autoregressive LLMs), 
we define attention sinks dynamically based on incoming attention mass.
For each attention head $h$ at layer $\ell$ and timestep $t$, we compute the 
\emph{incoming attention mass} for each key position $j$:
\begin{equation}
    m_j^{(\ell,t,h)} = \frac{1}{N} \sum_{i=1}^{N} A_{i,j}^{(\ell,t,h)},
\end{equation}
where $A_{i,j}^{(\ell,t,h)}$ denotes the attention weight from query $i$ to key $j$.
The \emph{dynamic top-$k$ sinks} are then defined as:
\begin{equation}
    S^{(\ell,t,h)} = \mathrm{TopK}\bigl(m^{(\ell,t,h)}, k\bigr).
\end{equation}
This definition identifies dominant attention recipients on a per-head, per-timestep basis 
without assuming any fixed token position.
We note that $k$ is defined per head and per timestep; however, the effective number of 
unique masked tokens is substantially smaller than $H \times k \times T$, as multiple 
heads frequently select the same dominant text tokens across the sequence.
We quantify sink strength using the \emph{maximum incoming mass}:
\begin{equation}
    \text{MaxMass}^{(\ell,t)} = \frac{1}{H} \sum_{h=1}^{H} \max_j m_j^{(\ell,t,h)}.
\end{equation}

Table~\ref{tab:dynamic_sink} reports CLIP-T results for three conditions: single-layer (L12), multi-layer (L6+12+18), and stronger intervention (top-5).
For the single-layer and multi-layer conditions, all 95\% CIs include zero, confirming that removing dynamically identified sinks does not degrade quality.

The stronger intervention (top-5) yields a small positive shift in CLIP-T ($\Delta = +0.0011$, $p = 0.01$); this effect remains within our predefined practical equivalence margin ($|\Delta| < 0.002$). We evaluate this condition further with HPS-v2 below.

\begin{figure*}[t]
\centering
\includegraphics[width=\textwidth, height=0.4\textheight, keepaspectratio]{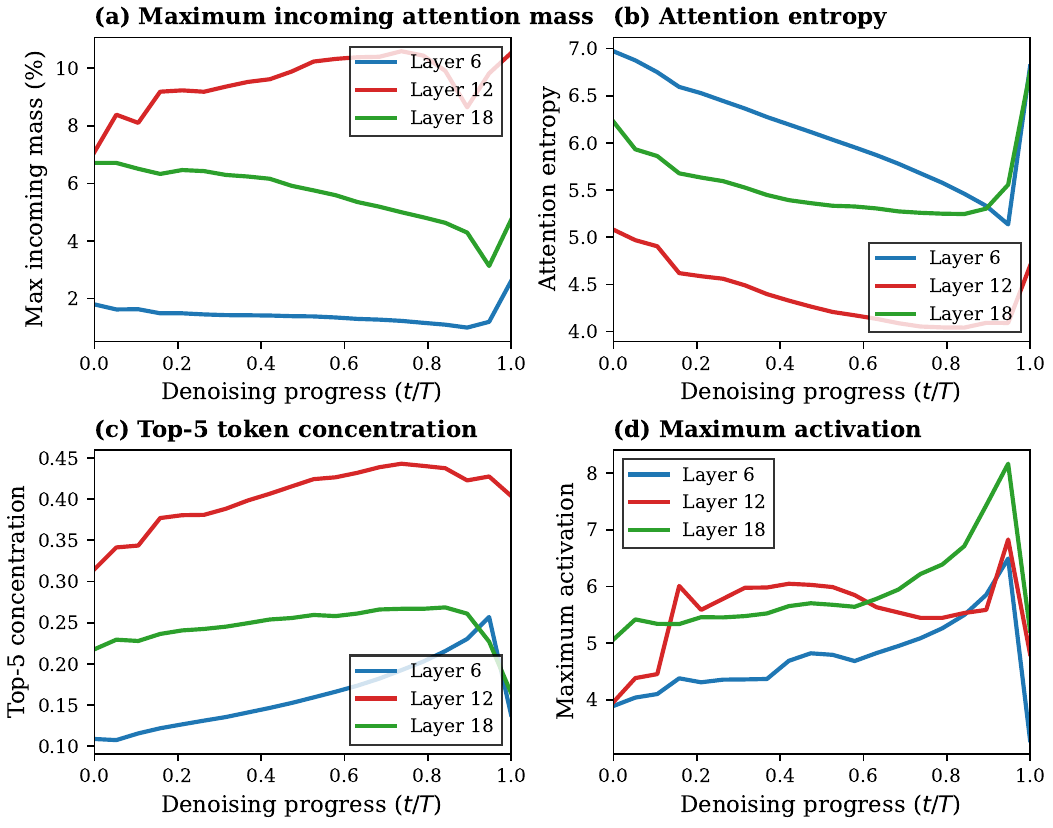}
\caption{\textbf{Attention sink dynamics across layers and denoising timesteps.} (a) Maximum incoming mass peaks at Layer~12 and decreases during denoising. (b) Attention entropy is inversely correlated with attention concentration (MaxMass). (c) Top-5 concentration increases over time. (d) Maximum activation follows a similar pattern to MaxMass.}
\label{fig:h1_dynamics}
\end{figure*}

\begin{table}[t]
\centering
\caption{\textbf{Attention concentration statistics (GenEval, $N=553$).} $M_{mass}$: Max attention mass; $T_5$: Top-5 concentration; $A_{idx}$: Max activation index; $O_{idx0}$: Overlap with index-0. Middle layers exhibit peak concentration while remaining distinct from index-0.}
\label{tab:h1_stats}
\vskip 0.1in
\begin{small}
\setlength{\tabcolsep}{4pt}
\renewcommand{\arraystretch}{1.08}
\begin{tabular}{lccccc}
\toprule
Layer & $M_{mass}$ & Entropy & $T_5$ Conc. & $A_{idx}$ & $O_{idx0}$ \\
\midrule
6 (shallow) & 1.9\% & 6.5--7.0 & 0.11 $\to$ 0.24 & 7--10 & $<$0.2\% \\
\rowcolor{sinkblue}
\textbf{12 (middle)} & \textbf{9.5\%} & \textbf{4.0--4.5} & \textbf{0.31 $\to$ 0.42} & \textbf{15--21} & \textbf{$<$0.2\%} \\
18 (deep) & 5.1\% & 5.0--6.5 & 0.19 $\to$ 0.29 & 9--13 & $<$0.1\% \\
\bottomrule
\end{tabular}
\end{small}
\end{table}

Together, Table~\ref{tab:h1_stats} and Figure~\ref{fig:h1_dynamics} yield five findings:
\begin{enumerate}[nosep,leftmargin=*]
    \item \textbf{Layer-wise concentration}: Layer~12 (middle) exhibits the strongest attention concentration, with maximum incoming mass $\approx$10\%, lowest entropy (4.0--4.5), and highest activation magnitudes (15--21).
    \item \textbf{Phase-dependent dynamics}: Following the timestep ordering of the official pipeline, we define normalized time $t/T \in [0,1]$ such that $t/T \approx 0$ corresponds to the noisiest denoising step.
    Attention concentration peaks during early denoising and diminishes toward later steps.
    \item \textbf{Concentration--entropy anti-correlation}: Layers with stronger attention concentration exhibit lower entropy, indicating more peaked attention distributions.
    \item \textbf{Dynamic sink $\neq$ index-0}: Unlike autoregressive LLMs where attention sinks typically coincide with special tokens (e.g., BOS), the dynamically identified dominant attention recipients in SD3 are \emph{not} at index-0. The overlap between dynamic top-1 sinks and index-0 is $<0.1\%$ across all layers and timesteps.
    \item \textbf{Position locality}: Top-1 dynamic sinks consistently occupy a narrow range of key indices ($\approx$4016--4231 in SD3's joint sequence) corresponding predominantly to early text-encoder positions. The position drift across (layer, $t/T$) is structural rather than free, suggesting that ``dynamic'' refers to localized shifts within a small subset of conditioning positions rather than movement across the full key space. A token-identity decomposition of which specific text tokens these sinks correspond to (e.g., padding, structural tokens, or content-bearing tokens), in the spirit of the register-token analysis of \citet{darcet2024registers} and \citet{jiang2025registers}, is left to future work.
\end{enumerate}

These observations suggest that \textbf{attention concentration and dominant recipients}
are transient and phase-dependent, emerging most prominently
during the high-noise regime of early denoising.

\subsection{H2: Causal Necessity of Attention Sinks}
\label{sec:h2}

Having established that sinks exist, we now test whether they are 
\emph{functionally necessary} for generation quality via causal interventions.

\textbf{Methodological note on ``causal.''}
We use ``causal'' in the interventionist sense~\citep{pearl2009causality}: 
paired, training-free manipulations that directly alter internal computations 
while holding all other factors (prompt, seed, hyperparameters) fixed.
This design isolates the effect of sink removal from confounding variation.
While such ablations do not establish necessity for \emph{all possible} 
downstream capabilities, they are sufficient to falsify necessity claims 
for the evaluated outcomes (alignment and preference metrics).

\subsubsection{Dynamic Sink Interventions}
\label{sec:dynamic_sink}

We conduct causal interventions using our dynamic sink definition, which identifies the top-$k$ positions with highest incoming attention mass on a per-head, per-timestep basis.
For each attention head at the target layer(s), we compute the incoming mass distribution and identify the top-$k$ positions dynamically.
We then apply a hard mask by subtracting $10^4$ from the pre-softmax logits of these positions, effectively zeroing their attention weights.
Table~\ref{tab:dynamic_sink} reports results for three conditions: single-layer (L12), multi-layer (L6+12+18), and stronger intervention (top-5).

\begin{table}[t]
\centering
\caption{\textbf{Dynamic sink intervention results (GenEval, $N=553$).} All metric changes ($\Delta_C$) and 95\% CIs are scaled by $10^3$. A value of $1.0$ corresponds to a 0.001 absolute change in CLIP-T score. All primary shifts are negligible ($|\Delta_C| < 2.0$).}
\label{tab:dynamic_sink}
\vskip 0.1in
\begin{small}
\addtolength{\tabcolsep}{0pt} 
\begin{tabular}{lcccrc}
\toprule
Condition & Layers & $k$ & $\Delta_{C}$ ($\times 10^{-3}$) [95\% CI] & $p$ \\
\midrule
Single-L & 12 & 1 & $-0.5$ [$-1.7, +0.7$] & 0.44 \\
Multi-L & 6, 12, 18 & 1 & $+0.9$ [$-0.1, +2.0$] & 0.08 \\
Top-5 & 12 & 5 & $+1.1$ [$+0.3, +2.0$] & 0.01 \\
\bottomrule
\end{tabular}
\end{small}
\vspace{-4ex}
\end{table}

\textbf{Evaluation with a stronger preference model.}
To mitigate concerns that CLIP-based metrics may miss subtle degradations,
we additionally evaluate dynamic sink interventions using HPS-v2,
a learned preference model trained on human preference data (Table~\ref{tab:hpsv2}).

\begin{table}[t]
\centering
\caption{\textbf{HPS-v2 evaluation of dynamic sink interventions (GenEval, $N=553$).}
A1 and A2 show no significant change; A3 exhibits a small negative shift.}
\label{tab:hpsv2}
\vskip 0.1in
\begin{small}
\begin{tabular}{lcccc}
\toprule
Condition & Layers & $k$ & $\Delta$HPS-v2 & $p$ \\
\midrule
A1 (Single-L) & 12 & 1 & $+0.0003$ & 0.42 \\
A2 (Multi-L) & 6, 12, 18 & 1 & $+0.0001$ & 0.85 \\
A3 (Top-5) & 12 & 5 & $-0.0020$ & 0.001 \\
\bottomrule
\end{tabular}
\end{small}
\end{table}

For the primary conditions (A1: single-layer top-1; A2: multi-layer top-1),
HPS-v2 shows no statistically significant change, with mean shifts near zero
and 95\% confidence intervals covering zero, consistent with CLIP-T and ImageReward.

Under the stronger A3 intervention (top-5 removal), HPS-v2 exhibits a small
negative shift ($\Delta=-0.0020$, $p=0.001$), while CLIP-T shows a small positive
shift ($\Delta=+0.0011$). Importantly, both effects are below 1\% in magnitude
and lie near the boundary of practical equivalence, indicating a marginal
metric-dependent trade-off rather than a meaningful quality change.

The intervention successfully eliminates attention to dynamically identified sinks.
At Layer~12, the top-1 incoming mass is reduced from 10.0\% to $<$0.001\%, achieving reduction factors exceeding $10^8\times$ across all tested layers (Table~\ref{tab:dynamic_verification}).

\begin{table}[t]
\centering
\caption{\textbf{Dynamic sink intervention verification (GenEval).} 
Reduction factors confirm effective removal of dominant attention recipients.}
\label{tab:dynamic_verification}
\begin{tabular}{lccc}
\toprule
Layer & Before & After & Reduction \\
\midrule
6  & 1.90\% & $<$0.001\% & $1.9 \times 10^8$ \\
\textbf{12} & \textbf{9.47\%} & $<$\textbf{0.001\%} & $\mathbf{9.5 \times 10^8}$ \\
18 & 5.15\% & $<$0.001\% & $5.1 \times 10^8$ \\
\bottomrule
\end{tabular}
\vspace{-4ex}
\end{table}

These results significantly strengthen our main conclusion: even when sinks are identified dynamically as the true dominant attention recipients (rather than a fixed proxy position), their removal produces no measurable quality degradation.

Our interventions test \emph{aggregation-level} necessity---whether sink
tokens must contribute their value vectors for the evaluated outcomes.
We do not test \emph{encoding-level} necessity (whether sinks must
contribute to key/query projections); doing so would require modifying
upstream layers and is outside our scope.

\subsection{Dose--Response Analysis}
\label{sec:dose_response}

To rule out threshold effects, we perform full dose--response sweeps 
across both score-path ($\eta \in [0, 1]$) and value-path interventions.
Both sweeps yield uniformly flat curves with all 95\% CIs covering zero, 
even at complete removal ($\eta = 0$, reduction $>44{,}000\times$).
Full results are in Appendix~\ref{sec:dose_response_appendix}.

\subsection{Robustness and Generalization}
\label{sec:robustness}

We test robustness across intervention scope, timing, and architecture.
Full results are in Appendix~\ref{sec:extended_robustness}.

Simultaneous intervention on layers 6, 12, and 18 produces no degradation
($\Delta$CLIP-T $= +0.0012$, $p = 0.68$).
Phase-specific interventions (early/mid/late only) likewise show no 
quality loss; notably, early-phase intervention when sink strength 
is maximal produces a slight \emph{improvement} ($\Delta = +0.0006$), 
contradicting the hypothesis that sinks serve a critical anchoring function.

\subsubsection{Cross-Architecture Validation: SDXL}

To test generalization beyond SD3's joint attention, we intervene on
both mid-block attention modules of SDXL: \emph{self-attention}
(\texttt{attn1}, image--image) and \emph{cross-attention}
(\texttt{attn2}, image--text, where keys/values are text embeddings).
We focus on the mid-block as it concentrates the strongest text--image
coupling; extending to early or late blocks is left for future work.
Sanity checks confirm: (1)~no-op produces pixel-identical outputs, and
(2)~intervention reduces sink mass by $>10^9\times$.

\begin{table}[t]
\centering
\caption{\textbf{SDXL cross-architecture validation ($N=100$, paired).}
$\Delta$CLIP-T is the paired change in CLIP-T. Both modules yield
95\% CIs containing zero; cross-attention induces larger perceptual
drift, consistent with the SD3 attribution result that text-side
sinks dominate trajectory perturbation
(Section~\ref{sec:robustness_attribution}).}
\label{tab:sdxl}
\vskip 0.1in
\begin{small}
\resizebox{0.48\textwidth}{!}{%
\begin{tabular}{lcc}
\toprule
SDXL Mid-Block & $\Delta$CLIP-T [95\% CI] & LPIPS \\
\midrule
Self-attention (\texttt{attn1})  & $+0.0004$ [$-0.0005, +0.0014$] & 0.045 \\
Cross-attention (\texttt{attn2}) & $-0.0003$ [$-0.0019, +0.0012$] & 0.077 \\
\bottomrule
\end{tabular}}
\end{small}
\end{table}

Both modules yield CIs containing zero. Cross-attention induces larger
perceptual drift (LPIPS 0.077 vs.\ 0.045), consistent with our SD3 E3
observation that text-side sinks produce larger trajectory perturbations
(Section~\ref{sec:robustness_attribution}).

\subsubsection{Perceptual and Distributional Effects}
\label{sec:perceptual}

Beyond alignment and preference metrics, we quantify how sink removal 
changes generated images relative to baseline outputs using paired 
perceptual distance (LPIPS) and same-domain distributional distance 
(FID$_{\text{shift}}$ computed between intervention and baseline outputs, 
not against real images).

Table~\ref{tab:perceptual} reports results across conditions. 
Perceptual and distributional shifts increase monotonically with 
intervention intensity: A1 (single-layer, top-1) induces moderate 
change (LPIPS $\approx 0.19$), while A3 (top-5) produces larger shifts 
(LPIPS $\approx 0.31$). Importantly, these shifts occur \emph{without} 
degrading CLIP-T, ImageReward, or HPS-v2, indicating that suppressing 
dominant attention recipients moves samples within the model's output 
manifold while preserving standard alignment and preference scores.

To confirm that observed shifts arise from the intervention itself
rather than implementation artifacts, we conduct a no-op sanity check: 
installing our attention processor with \texttt{intervention\_enabled=False}. 
Across 100 paired generations, outputs are pixel-identical to unmodified 
baseline (pixel diff = 0, LPIPS = 0, FID$_{\text{shift}}$ = 0), 
confirming that the substantial perceptual changes under active 
intervention arise solely from attention sink removal 
(see Appendix~\ref{sec:sanity} for details).

\begin{table}[t]
\centering
\caption{\textbf{Perceptual and distributional effects.} 
LPIPS measures paired perceptual distance; FID$_{\text{shift}}$ measures 
distributional change between intervention and baseline outputs. 
Semantic alignment (CLIP-T) is preserved while perceptual similarity 
decreases with intervention intensity.}
\label{tab:perceptual}
\vskip 0.1in
\begin{small}
\begin{tabular}{llccccc}
\toprule
Dataset & Cond. & $N$ & $\Delta$CLIP-T & LPIPS & FID$_{\text{shift}}$ \\
\midrule
\multirow{3}{*}{GenEval} 
  & A1 & 553 & \nullcell{$+0.001$} & \driftcell{0.189} & \driftcell{432} \\
  & A2 & 553 & \nullcell{$+0.001$} & \driftcell{0.242} & \driftcell{992} \\
  & A3 & 553 & \nullcell{$-0.001$} & \driftcell{0.314} & \driftcell{996} \\
\midrule
COCO-5k & A1 & 5000 & $-0.000$ & 0.189 & 727 \\
\midrule
\textit{No-op} & -- & 100 & -- & 0.000 & 0 \\
\bottomrule
\end{tabular}
\end{small}
\vskip -0.1in
\end{table}

\subsubsection{Sink-Specific Perceptual Dissociation}
\label{sec:sink_specificity_main}

To verify that the perceptual shifts induced by sink suppression are
not an artifact of removing arbitrary tokens, we compare sink masking
against \emph{equal-budget random masking} using a paired
difference-of-differences statistic
$\Delta\Delta = \mathrm{LPIPS}_{\text{sink}} - \mathrm{LPIPS}_{\text{rand}}$.
These comparisons use the union-budget protocol described below
(layer~12, $N{=}64$).

We distinguish two masking protocols: (i)~\emph{per-head top-$k$},
used in the dynamic-sink interventions of
Section~\ref{sec:dynamic_sink}, where each head independently masks
its own top-$k$ keys; and (ii)~\emph{union-budget top-$k$}, used in
this section, where the top-$k$ keys ranked by head-averaged incoming
mass are masked uniformly across all heads. The latter is more
aggressive and is used here to enable matched-budget comparison
against random masking.

\begin{table}[t]
\centering
\caption{\textbf{Sink vs.\ random masking at equal budget
(union-budget protocol, layer~12, $N{=}64$).}
$\Delta\Delta = \mathrm{LPIPS}_{\text{sink}} - \mathrm{LPIPS}_{\text{rand}}$.
Sink masking induces substantially larger perceptual drift than
equal-budget random masking at both budgets.}
\label{tab:sink_vs_random_lpips}
\vskip 0.1in
\begin{small}
\resizebox{0.48\textwidth}{!}{%
\begin{tabular}{cccccc}
\toprule
$k$ & LPIPS$_{\text{sink}}$ & LPIPS$_{\text{rand}}$ & $\Delta\Delta$ & 95\% CI & $p$ \\
\midrule
1 & 0.347 & 0.053 & $+0.295$ & $[+0.265, +0.323]$ & $<0.0001$ \\
5 & 0.436 & 0.104 & $+0.332$ & $[+0.308, +0.358]$ & $<0.0001$ \\
\bottomrule
\end{tabular}}
\end{small}
\end{table}

Sink masking induces $\sim\!6\times$ larger perceptual shift than
equal-budget random masking at $k{=}1$, with the gap widening at
$k{=}5$. This dissociation---strong trajectory perturbation alongside
preserved alignment---suggests that sinks carry structured
trajectory-level information while remaining unnecessary for alignment.
Representative qualitative comparisons that visualize this dissociation 
are provided in Appendix~\ref{sec:qualitative_panels}.

\subsection{Robustness and Attribution Analyses}
\label{sec:robustness_attribution}

We conduct additional analyses to assess task-type robustness (E1), 
sampling sensitivity across CFG/steps/schedulers (E2), and text vs.\ 
image sink attribution under SD3's joint attention (E3).
All analyses confirm that $\Delta$CLIP-T CIs include zero across conditions.
Notably, under SD3 joint attention, over 99.9\% of dynamically identified sinks
correspond to text-conditioning tokens rather than visual-latent tokens
(47,999/48,000 records), suggesting that dominant attention recipients are
concentrated on the text modality.
Full results are in Appendix~\ref{sec:attribution_appendix}.

We further validate alignment robustness with BLIP2-VQA~\citep{li2023blip2},
a non-CLIP compositional probe. On the full $N{=}553$ image set used
in Table~\ref{tab:dynamic_sink} (per-head protocol), BLIP2-VQA shows
no detected sink-specific effect (paired $\Delta = +0.0001$,
95\% CI $[-0.0039, +0.0040]$, $p = 0.97$); a smaller-$N$ rebuttal
subset ($N{=}64$) likewise yields
$\Delta\Delta = -0.0074$, 95\% CI $[-0.0215, +0.0056]$, $p = 0.27$.
This convergent evidence outside the CLIP embedding space supports
the alignment-preservation finding.

On a 24-prompt compositional subset (color binding, spatial, counting),
all three CLIP-decomposed sub-concept scores produce 95\% CIs containing
zero; given the small $N$ and CLIP-derivative scoring, we view this as
supporting evidence rather than a replacement for dedicated benchmarks.

\subsection{Summary of Findings}
\label{sec:summary}

Table~\ref{tab:summary} consolidates our experimental findings. Figure~\ref{fig:summary_heatmap} in Appendix provides a visual summary of all experimental results.

\begin{table*}[t]
\centering
\small
\setlength{\tabcolsep}{3pt}
\caption{\textbf{Summary of experimental findings.}}
\label{tab:summary}
\begin{tabular}{lll}
\toprule
Question & Answer & Evidence \\
\midrule
Do dominant attn.\ recipients exist? 
& \sinkcell{Yes} 
& Dynamic top-1 mass $\approx$9.5\% (L12) \\
Are they at a fixed position? 
& \sinkcell{No} 
& Index-0 overlap $<$0.2\% \\
Are they phase-dependent? & Yes & Peaks early, diminishes late \\
Does removal hurt quality? 
& \nullcell{\textbf{No}} 
& All $|\Delta| < 0.002$, primary CIs $\ni 0$ \\
Does stronger removal hurt? & \textbf{No} & Top-5: within margin (both metrics) \\
Is this metric-consistent? & Yes & CLIP-T, IR, HPS-v2 agree (A1/A2) \\
Does multi-layer removal hurt? & \textbf{No} & L6+12+18: $\Delta = +0.0009$ \\
Is it arch-general? & Yes & SD3 + SDXL consistent \\
Does it change perceptual similarity? 
& \driftcell{Yes} 
& LPIPS: 0.19 (A1) $\to$ 0.31 (A3) \\
Is the change from intervention? 
& \ctrlcell{\textbf{Yes}} 
& No-op sanity: LPIPS = 0 \\
\bottomrule
\end{tabular}
\end{table*}

\section{Discussion}

Our experiments provide strong evidence that attention sinks in diffusion 
transformers are not necessary for semantic alignment and do not affect 
preference metrics under standard inference settings ($k{=}1$).

\textbf{Interpreting perceptual and distributional shifts.}
The observed LPIPS and FID$_{\text{shift}}$ changes should not be interpreted 
as quality degradations.
Critically, FID$_{\text{shift}}$ measures distributional distance 
\emph{between baseline and intervention outputs}, not against real images.
To contextualize these values, Table~\ref{tab:fid_context} reports 
FID under common variations without any attention intervention.
\begin{table}[t]
\centering
\caption{\textbf{FID calibration baselines.} FID$_{\text{shift}}$ from 
common variations provides reference points for interpreting intervention effects.}
\label{tab:fid_context}
\vskip 0.1in
\begin{small}
\begin{tabular}{lc}
\toprule
Variation & FID \\
\midrule
Seed variation (same settings) & 115 \\
CFG $\pm 1$ & 54--58 \\
Steps $-5$ to $-10$ & 81--109 \\
Scheduler change & 331 \\
\midrule
\textbf{Sink intervention (ours)} & \textbf{432--996} \\
\bottomrule
\end{tabular}
\end{small}
\end{table}
The intervention-induced FID$_{\text{shift}}$ (400--1000) is comparable 
in magnitude to aggressive scheduler substitutions, rather than indicating 
anomalous distributional collapse.
This calibration confirms that sink suppression alters visual appearance by shifting samples within the model’s output manifold, yet does so without compromising alignment or human preference metrics.
For applications focused on ranking, alignment evaluation, or preference 
optimization, these shifts are immaterial.
For applications requiring perceptual consistency (e.g., video generation, 
image editing), sink suppression may warrant additional constraints.

Under standard settings ($k{=}1$), reducing sink attention by $>10^8\times$
produces no degradation under CLIP-T, ImageReward, or HPS-v2. CLIP-T shows 
a slight positive shift under both sink and random removal, indicating that 
single-token masking does not harm (and may marginally benefit) the alignment.
Under stronger top-5 removal, metrics show small ($<$1\%) divergent shifts 
(CLIP-T slightly positive, HPS-v2 slightly negative), indicating 
metric-dependent variability rather than meaningful quality change.
Unlike AR models where sinks coincide with BOS tokens, diffusion 
transformers exhibit no fixed positional sink (index-0 overlap $<$0.2\%).
Non-functionality holds across intervention types, strengths, layers, 
phases, and architectures. Our results do not contradict prior 
attention-editing or guidance methods: usefulness for optimization 
or controllability does not imply necessity for semantic alignment.
We emphasize that our claims concern alignment as measured by widely 
used proxy metrics (CLIP-T, HPS-v2); subtle compositional errors 
may require detector-based or human evaluation and remain an important 
direction for future work.

\textbf{Perceptual shifts as a boundary condition.}
Suppressing sinks induces LPIPS $\approx$0.06--0.31 and FID$_{\text{shift}}$
$\approx$400--1000 (comparable to seed variation or CFG changes;
see Appendix~\ref{sec:fid_calibration}).
These shifts scale with intervention intensity and arise solely from
the intervention (no-op sanity: LPIPS $= 0$).
In SD3 joint attention, text-sink suppression induces larger shifts
(LPIPS $\approx 0.16$) than in SDXL cross-attention (LPIPS $\approx 0.06$),
consistent with tighter text--image coupling in joint-attention architectures.
Together, these results reveal an empirical dissociation: sink suppression
strongly perturbs trajectory-level realization (LPIPS $\sim\!6\times$
larger than equal-budget random masking;
Section~\ref{sec:sink_specificity_main}) while leaving proxy-level
semantic alignment unchanged. Sinks thus appear to carry structured
trajectory information without being necessary for alignment.

\textbf{Reconciling with theoretical sink necessity.}
Recent theoretical work~\citep{trigger2026} establishes that softmax attention 
sinks are necessary for trigger-conditional tasks, where models must implement 
a stable no-op state under normalization constraints. Our findings do not 
contradict this: we test \emph{semantic-alignment necessity} in diffusion image 
generation, a regime where the trigger-conditional analysis does not directly 
apply. Whether the alignment-irrelevance of sinks observed here reflects a 
fundamentally different functional role of sinks in non-autoregressive bidirectional 
attention, or simply a task that does not exercise the trigger-conditional 
machinery, remains an open question.

\textbf{Preference trade-off under stronger interventions.}
To assess sink-specificity, we compare sink masking against equal-budget 
random masking using a paired difference-of-differences test: 
$\Delta\Delta = \Delta_{\text{sink}} - \Delta_{\text{rand}}$.
Under standard settings ($k{=}1$), neither CLIP-T nor HPS-v2 shows a 
sink-specific effect ($\Delta\Delta$ CIs include zero for both metrics; 
Table~\ref{tab:sink_specificity_appendix}). Both sink and random single-token 
removal are tolerated, confirming that the non-necessity finding is not 
an artifact of removing ``any'' token.
However, stronger interventions ($k \geq 10$) reveal a metric-dependent boundary.
At masking budgets $k \in \{10, 50\}$, HPS-v2 shows sink-specific degradation.
Under HPS-v2, sink masking degrades preference scores significantly more 
than random masking at both $k{=}10$ ($\Delta\Delta = -0.005$, 95\% CI: 
$[-0.008, -0.001]$, one-sided $p = 0.007$) and $k{=}50$ ($\Delta\Delta = -0.020$, 
CI: $[-0.026, -0.013]$, $p < 10^{-4}$). A paired trend test confirms that 
this effect increases across tested budgets ($k{=}10, 50$)
($\Delta d = -0.015$, CI: $[-0.022, -0.008]$, $p < 10^{-4}$).
Importantly, CLIP-T remains stable across all tested budgets (Table~\ref{tab:clip_k_sweep};
$k \in \{1, 5, 10, 20, 50\}$, all CIs include zero), confirming that
alignment is robust regardless of intervention intensity. This metric-dependent 
pattern suggests that sinks may contribute to preference-oriented quality 
dimensions not captured by alignment proxies. We note that HPS-v2 is a 
learned preference proxy rather than a human study; we interpret these 
effects as indicative of preference-oriented dimensions, not definitive 
perceptual judgments. Full results are in Appendix~\ref{sec:sink_specificity}.

\textbf{Sink definition and modality imbalance.}
We note that the dynamic sink statistic aggregates incoming mass over queries, 
which naturally elevates text keys given the larger number of visual queries. 
Our conclusions do not rely on sink modality; rather, they concern the functional 
necessity of dominant recipients as identified by this operational definition. 
We treat sink modality imbalance as a property of the architecture, 
not as a confound for the non-necessity claim.

\textbf{Implications and limitations.}
Attention sinks can be safely suppressed, potentially enabling sparse
attention patterns or efficiency improvements.
We use GenEval primarily as a large, diverse prompt set to ensure statistical
power for paired causal comparisons, rather than as a detector-based benchmark;
our non-necessity claim is therefore scoped to semantic alignment proxies
(CLIP-T) and preference metrics (HPS-v2), not fine-grained compositional
correctness.
Detector-based compositional benchmarks (e.g., T2I-CompBench~\citep{huang2023t2icompbench},
Gecko~\citep{gecko2024}) remain a natural follow-up for stress-testing
alignment claims at sub-concept granularity.
We do not provide realized speedups, latency gains, or FLOPs reductions;
our contribution is the causal prerequisite---dominant recipients can be
suppressed without alignment loss---rather than a system-level efficiency
result.
This study does not evaluate perceptual fidelity via human studies or
demonstrate end-to-end efficiency gains---directions that are complementary
but orthogonal to our central question.
Sink-specific trajectory control may also enable applications such as
generation steering, an intriguing direction we leave open.

\textbf{Design implications and engineering caveats.}
A concrete implication of these findings is that sparsification schemes
for diffusion transformers need not hard-code dominant attention
recipients as privileged tokens to be preserved at inference time;
instead, they can be treated as candidates for budgeted removal while
monitoring the target metric. Translating this into wall-clock speedups
requires further engineering (dynamic token-selection overhead, fused
sparse kernels) that we leave to future work.

\section{Conclusion}
We presented a causal analysis of attention sinks in diffusion transformers, 
defining sinks dynamically as key positions receiving dominant incoming 
attention mass and intervening along both the score and value paths. 
Across large-scale paired evaluations on SD3 and SDXL, suppressing 
dynamically identified sinks does not degrade semantic alignment or 
preference metrics under standard inference settings ($k{=}1$), yet 
induces sink-specific perceptual shifts---roughly $6\times$ larger than 
equal-budget random masking---that mark a structural boundary condition. 
Under stronger interventions ($k\!\geq\!10$), preference proxies (HPS-v2) 
exhibit a metric-dependent degradation that increases with intervention 
intensity, while alignment (CLIP-T) remains robust throughout. These 
findings caution against transferring autoregressive sink intuitions to 
diffusion transformers, and suggest that high incoming attention mass 
need not indicate functional necessity for alignment in non-autoregressive 
generation.

\section*{Acknowledgements}
Research reported in this publication was supported by DOE ASCR under Award Number DE-SC0022873, the National Institutes of Health under Award Number R01GM143789, and the Advanced Research Projects Agency for Health (ARPA-H) under Award Number D24AC00338-00. The content is solely the responsibility of the authors and does not necessarily represent the official views of the National Institutes of Health, the Department of Energy, or the Advanced Research Projects Agency for Health.

\section*{Impact Statement}

This paper presents work whose goal is to advance the field of Machine 
Learning. 
There are many potential societal
consequences of our work, none which we feel must be
specifically highlighted here.

\bibliography{example_paper}
\bibliographystyle{icml2026}

\newpage
\appendix
\onecolumn
\section{Experimental Details}

\subsection{Models}

We conduct experiments primarily on \textbf{Stable Diffusion 3 (SD3)} using the official inference pipeline. In this pipeline, attention is computed over a \emph{mixed token set} containing both visual-latent tokens and text-conditioning tokens within the same attention computation (i.e., attention is not purely cross-attention over a disjoint context). This setting is particularly relevant because it exposes attention-sink behavior under a joint-attention regime.

To validate cross-architecture generality, we additionally evaluate on \textbf{Stable Diffusion XL (SDXL)}, a U-Net--based diffusion model with conventional self- and cross-attention blocks. For SDXL, sink interventions are applied to self-attention layers in the U-Net mid-block.

All models are used in inference mode with pretrained weights; \textbf{no finetuning} is performed.

\subsection{Prompts and Sampling}

For our \textbf{primary analysis} (dynamic sink interventions, Table~\ref{tab:dynamic_sink}), we evaluate on \textbf{553 prompts} from the GenEval benchmark~\cite{geneval}, which provides a diverse set of object-centric, scene-centric, and compositional queries. This large sample size ensures high statistical power ($>$99\% for detecting $\Delta > 0.001$) and narrow confidence intervals (95\% CI width $\approx \pm 0.001$).

For \textbf{supplementary analyses} (dose--response sweeps, phase-specific interventions, cross-architecture validation), we use \textbf{32 prompts} selected to balance diversity and computational efficiency. These experiments aim to verify trends and robustness rather than establish precise effect sizes.

All experiments use \textbf{paired sampling}: for prompt index $i$, the random seed is fixed to $\text{seed}+i$ across all intervention conditions. This ensures that differences between conditions are attributable solely to the intervention rather than stochastic variation.

Unless otherwise stated, we use \textbf{20 diffusion steps} with the default scheduler and guidance settings provided by each official pipeline.

\subsection{Index-Based Sink Proxy (Baseline Definition)}
\label{sec:appendix_index_sink}

We initially considered a fixed index-based proxy for attention sinks, following observations in autoregressive language models where the first token (e.g., BOS) often receives disproportionate attention~\citep{xiao2024streamingllm,sun2024massiveactivations,gu2024attention}.
Specifically, we designated the sink as the first token position in the attention sequence (index~0).

However, as shown in Section~\ref{sec:h1}, this position does \emph{not} correspond to the dominant attention recipients in SD3: the overlap between index-0 and the dynamically identified top-1 sink is $<$0.1\% across all layers and timesteps.
We therefore treat this fixed-position proxy as a \textbf{baseline or negative control} and adopt a dynamic sink definition (Section~\ref{sec:dynamic_sink}) as the primary analysis target.

For completeness, we report that interventions on index-0 also produce 
no quality degradation, which is consistent with (but weaker than) the 
dynamic sink results, since the fixed proxy does not target the true 
dominant attention recipients.

\subsection{H1: Sink Dynamics Measurement}

To characterize sink dynamics, we instrument attention blocks to record the following statistics at each diffusion timestep:
\begin{itemize}
    \item \textbf{Maximum incoming mass (MaxMass)}: for each head $h$ and timestep $t$, we compute the incoming
    attention mass for each key position $j$ as
    \[
    m_j^{(\ell,t,h)} = \frac{1}{N}\sum_{i=1}^{N} A_{i,j}^{(\ell,t,h)},
    \]
    and define $\mathrm{MaxMass}^{(\ell,t)} = \frac{1}{H}\sum_{h=1}^{H}\max_j m_j^{(\ell,t,h)}$.
    \item \textbf{Attention Entropy}:
    \[
    H(p) = -\sum_j p_j \log p_j,
    \]
    computed per query after clamping probabilities ($p_j \leftarrow \max(p_j, 10^{-12})$) for numerical stability.
    \item \textbf{Top-$k$ Concentration}: cumulative attention mass of the top-5 attended tokens.
    \item \textbf{Activation Magnitude}: maximum and 95th-percentile activation norms at the attention block output.
\end{itemize}
Statistics are aggregated over three representative layers (early, middle, late) corresponding to transformer layers 6, 12, and 18 in SD3. Reported curves are averaged over prompts.

\subsection{H2: Causal Sink Interventions}

We perform causal interventions on attention sinks along two independent pathways: the \textbf{score (logit) path} and the \textbf{value path}. All interventions are applied during inference.

\subsubsection{Score-path Intervention}

To suppress sink attention while preserving relative attention among non-sink tokens, we apply a \textbf{logit-bias intervention}:
\[
\ell_{\text{sink}} \leftarrow \ell_{\text{sink}} + \log(\eta),
\]
where $\eta \in [0,1]$ controls the strength of suppression. This operation is equivalent to scaling the sink probability by $\eta$ followed by renormalization, without injecting a uniform prior over non-sink tokens.

For $\eta=0$, we approximate complete suppression by adding a large negative bias (e.g., $-10^4$) to the sink logit before softmax. In practice, we apply standard numerical safeguards (clamping) to avoid overflow and observe \textbf{no NaNs} in any run.

\subsubsection{Value-path Intervention}

To test whether sink token content is semantically relevant, we modify the sink token's value vector using:
\begin{itemize}
    \item \textbf{Zero}: replace with zeros.
    \item \textbf{Mean}: replace with the mean value across tokens.
    \item \textbf{Lerp}: linear interpolation between original and mean values.
\end{itemize}
These modifications operate directly on the forward activations during inference.

\subsection{Dose--Response Sweeps}

To rule out threshold effects, we conduct full \textbf{dose--response sweeps}:
\begin{itemize}
    \item \textbf{Score sweep}: $\eta \in \{1.0, 0.5, 0.25, 0.1, 0.01, 0.0\}$.
    \item \textbf{Value sweep}: baseline, lerp$_{0.5}$, lerp$_{0.0}$, mean, zero.
\end{itemize}
For each condition, images are generated using paired seeds, and quality differences are computed relative to baseline.

We deem a response curve \textbf{practically flat} if all 95\% confidence intervals include zero and $\max |\mathbb{E}[\Delta]| < 0.002$ for CLIP-T and $< 0.05$ for ImageReward.

\subsection{Robustness Experiments}

\paragraph{Multi-layer Interventions.}
To test whether sinks become important when removed across multiple layers, we simultaneously apply sink suppression to layers $\{6, 12, 18\}$. Results are compared against single-layer (layer 12 only) and baseline conditions.

\paragraph{Phase-specific Interventions.}
To test phase dependence, we enable sink suppression only during specific portions of the diffusion trajectory:
\begin{itemize}
    \item \textbf{Early}: $t/T \in [0, 0.2]$,
    \item \textbf{Middle}: $t/T \in [0.4, 0.6]$,
    \item \textbf{Late}: $t/T \in [0.8, 1.0]$.
\end{itemize}
We define normalized time $t/T \in [0,1]$ following the pipeline's timestep ordering, where $t/T \approx 0$ corresponds to the \textbf{noisiest step} in our implementation. A callback mechanism toggles the intervention based on the normalized timestep. We log the fraction of timesteps during which the intervention is active to verify correct phase targeting.

\paragraph{Cross-architecture Validation (SDXL).}
On SDXL, sink suppression is applied to both mid-block attention modules
(self-attention \texttt{attn1} and cross-attention \texttt{attn2}) using
the same logit-bias formulation as in SD3. Per-module results
($N=100$, paired) are reported in Table~\ref{tab:sdxl}; both modules
yield 95\% CIs containing zero, with cross-attention inducing larger
perceptual drift than self-attention (LPIPS 0.077 vs.\ 0.045).

\subsection{Quality Metrics}

We evaluate generation quality using two complementary metrics:
\begin{itemize}
    \item \textbf{CLIP-T}: cosine similarity between CLIP text and image embeddings.
    \item \textbf{ImageReward}: a learned human-preference reward model.
\end{itemize}
Using both metrics mitigates reliance on any single proxy.

\subsection{Statistical Analysis}

All reported quality differences are computed as \textbf{paired differences}:
\[
\Delta_i = s_i^{\text{cond}} - s_i^{\text{baseline}},
\]
where $i$ indexes prompts.

We report the mean $\Delta$, \textbf{95\% bootstrap confidence intervals} as our primary uncertainty estimate, and additionally report paired $t$-test $p$-values as a secondary check.

\subsection{Intervention Verification}

To verify that interventions are effective, we explicitly measure sink ratio before and after intervention. In the strongest condition ($\eta=0$), sink attention mass is reduced from approximately 4.4\% to 0.0\%, corresponding to a \textbf{44,000$\times$ reduction factor}. Unless stated otherwise, reported sink-ratio reductions are averaged across heads and prompts at the intervened layers and at timesteps where the intervention is active.
Note that this verification uses the fixed index-0 proxy for implementation sanity checking; dynamic sink interventions are verified via MaxMass reductions reported in Table~\ref{tab:dynamic_verification}.

\subsection{Computational Resources and Reproducibility}

Experiments are run on NVIDIA \textbf{A6000} GPUs. Most experiments complete within a few hours; extensive sweeps are enabled by lightweight paired inference and do not require training.

For reproducibility, all runs:
\begin{itemize}
    \item use fixed random seeds with per-prompt pairing,
    \item save prompts and configurations alongside generated images,
    \item record intervention parameters and metrics in JSON format.
\end{itemize}
This enables exact reproduction of all reported results.

\section{Extended Results and Analysis}
\label{sec:extended_results}

\subsection{Summary of All Intervention Effects}

\begin{figure*}[t]
\centering
\includegraphics[width=0.8\textwidth]{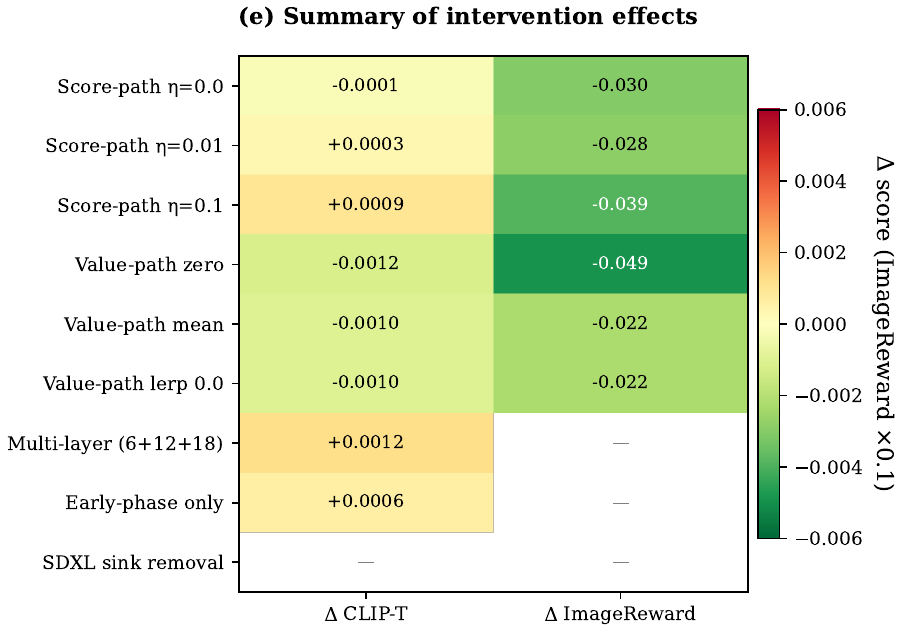}
\caption{\textbf{Summary of all experimental results.}
Heatmap showing paired $\Delta$ values across all interventions and metrics.
All effects are small and lie within a narrow range around zero; 
most are statistically non-significant, and all remain within the predefined practical equivalence margin ($|\Delta| < 0.002$ for CLIP-T), 
indicating no meaningful quality degradation from sink removal under any tested condition.}
\label{fig:summary_heatmap}
\end{figure*}

See Figure.~\ref{fig:summary_heatmap}.

\section{Experimental Setup}
\label{app:appendix_setup}

\paragraph{Model and Architecture.}
We conduct experiments primarily on Stable Diffusion~3 (SD3) using the official inference pipeline.
In this pipeline, attention is computed over a mixed token set containing both visual-latent tokens and text-conditioning tokens within the same attention operation (i.e., attention is not purely cross-attention).
As a result, all tokens in the joint sequence may potentially act as attention sinks.
This setting differs from conventional U-Net cross-attention and allows us to analyze attention-sink behavior under a more general transformer-style attention formulation.

For cross-architecture validation, we additionally evaluate on Stable Diffusion XL (SDXL), which uses a U-Net backbone with separate self-attention and cross-attention layers.
All models are used strictly in inference mode with pretrained weights; no finetuning is performed.

\paragraph{Evaluation Protocol.}
All experiments employ \textbf{strict paired generation}: for each evaluation prompt, we fix the random seed across all intervention conditions, ensuring that observed differences reflect intervention effects rather than sampling variance.
We report \textbf{mean paired differences} ($\Delta$) with \textbf{95\% bootstrap confidence intervals} (1,000 resamples). Statistical significance is assessed via paired $t$-tests.

For our primary analysis (dynamic sink interventions, Section~\ref{sec:dynamic_sink}), we use \textbf{553 prompts} from the GenEval benchmark~\cite{geneval}, providing high statistical power ($>$99\% for $\Delta > 0.001$).
Supplementary analyses (dose--response sweeps, phase-specific interventions) use 32 prompts; cross-architecture validation (SDXL) uses 100 prompts for tighter confidence intervals.

The practical equivalence margin ($|\Delta| < 0.002$ for CLIP-T) is chosen 
relative to the empirical noise floor observed under seed variation and 
bootstrap uncertainty at $N{=}553$, corresponding to changes below 
category-level sensitivity on GenEval.

\paragraph{Metrics.}
We evaluate generation outcomes using a suite of complementary metrics, 
grouped by their role in our analysis:

\begin{itemize}[nosep,leftmargin=*]
    \item \textbf{Alignment metrics (primary).}
    \textbf{CLIP-T} measures text--image semantic alignment via cosine similarity 
    between CLIP embeddings. This is our primary metric for testing 
    \emph{non-necessity} claims under standard settings.

    \item \textbf{Preference proxies (secondary).}
    \textbf{ImageReward} and \textbf{HPS-v2} are learned reward models trained 
    on human preference data, capturing perceptual quality and aesthetic appeal.
    These metrics are used to characterize boundary conditions under 
    stronger interventions.

    \item \textbf{Perceptual and distributional shift metrics (diagnostic).}
    \textbf{LPIPS} quantifies perceptual differences between baseline and 
    intervened outputs, while \textbf{FID$_{\text{shift}}$} measures 
    distributional changes between paired output sets. 
    These metrics diagnose trajectory changes rather than quality degradation.
\end{itemize}

This metric stratification clarifies which conclusions rely on which signals:
alignment robustness (CLIP-T), preference trade-offs (ImageReward, HPS-v2), 
and perceptual shifts (LPIPS, FID$_{\text{shift}}$).

\paragraph{Intervention Design.}
We intervene on attention sinks via two complementary pathways:
\begin{itemize}[nosep,leftmargin=*]
    \item \textbf{Score-path}: Scale the pre-softmax logit of sink tokens by adding $\log \eta$, which is equivalent to multiplying the post-softmax probability by $\eta$ (up to renormalization). Setting $\eta = 0$ corresponds to subtracting a large constant ($10^4$), effectively zeroing sink attention.
    \item \textbf{Value-path}: Replace the value vector at sink positions with alternatives (zero, mean of non-sink values, or linear interpolation).
\end{itemize}

\section{Dose--Response Full Results}
\label{sec:dose_response_appendix}

\begin{figure}[h]
\centering
\includegraphics[width=0.9\textwidth]{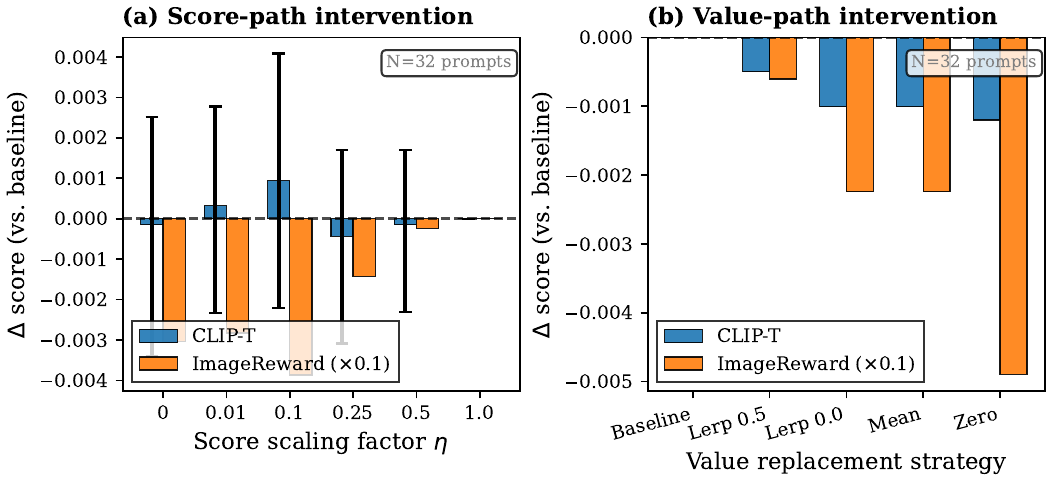}
\caption{\textbf{Dose--response curves for score-path (left) and value-path 
(right) interventions.} Error bars indicate 95\% bootstrap CIs. Both curves 
are flat across the entire intervention range, with all CIs covering zero.}
\label{fig:dose_response_appendix}
\end{figure}

This section provides full numerical results for the dose--response
analysis summarized in Section~\ref{sec:dose_response}; Figure~\ref{fig:dose_response_appendix} visualizes both score-path and value-path sweeps.

\subsection{Score-Path Sweep}

Tables~\ref{tab:score_sweep_clip_appendix} and~\ref{tab:score_sweep_ir_appendix} report the full score-path dose--response under CLIP-T and ImageReward respectively.

\begin{table}[h]
\centering
\caption{\textbf{Score-path dose--response (CLIP-T).} $\eta$: Attention suppression ratio ($\eta=0$ is full ablation). All metrics and 95\% CIs are scaled by $10^3$.}
\label{tab:score_sweep_clip_appendix}
\vskip 0.1in
\begin{small}
\begin{tabular}{ccc}
\toprule
Suppression $\eta$ & $\Delta_{C}$ ($10^{-3}$) [95\% CI] & $p$ \\
\midrule
0.5  & $-0.2$ [$-2.3, +1.7$] & 0.89 \\
0.25 & $-0.4$ [$-3.1, +1.7$] & 0.73 \\
0.1  & $+0.9$ [$-2.2, +4.1$] & 0.59 \\
0.01 & $+0.3$ [$-2.3, +2.8$] & 0.81 \\
\textbf{0.0 (Full)} & \textbf{$-0.1$ [$-3.4, +2.5$]} & \textbf{0.92} \\
\bottomrule
\end{tabular}
\end{small}
\end{table}

\begin{table}[h]
\centering
\caption{\textbf{Score-path dose--response (ImageReward).} Consistent with CLIP-T, all CIs include zero.}
\label{tab:score_sweep_ir_appendix}
\vskip 0.1in
\begin{small}
\begin{tabular}{ccc}
\toprule
Suppression $\eta$ & $\Delta_{IR}$ ($10^{-3}$) [95\% CI] & $p$ \\
\midrule
0.5  & $-2.4$ [$-32.4, +26.5$] & 0.88 \\
0.25 & $-14.2$ [$-75.4, +30.4$] & 0.63 \\
0.1  & $-38.7$ [$-134.5, +23.0$] & 0.39 \\
0.01 & $-28.3$ [$-125.8, +29.0$] & 0.51 \\
\textbf{0.0 (Full)} & \textbf{$-30.3$ [$-119.5, +29.1$]} & \textbf{0.48} \\
\bottomrule
\end{tabular}
\end{small}
\end{table}

\subsection{Intervention Verification}

\begin{figure}[h]
\centering
\includegraphics[width=0.6\textwidth]{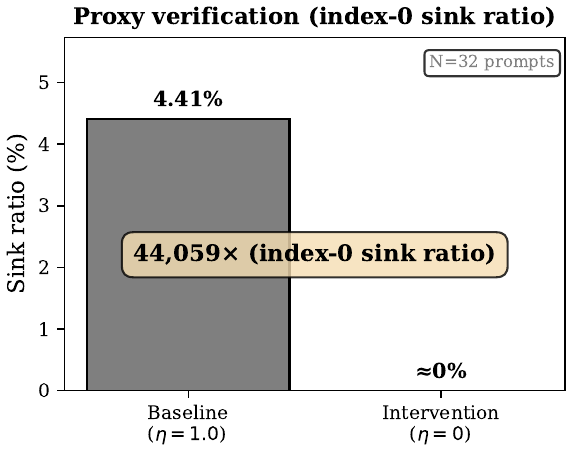}
\caption{\textbf{Intervention verification for the index-based sink proxy (index-0).} Complete removal ($\eta = 0$) reduces the index-0 sink ratio by 44,059$\times$, confirming effectiveness of the proxy intervention.}
\label{fig:verification_appendix}
\end{figure}

Figure~\ref{fig:verification_appendix} reports the index-0 sink ratio before and after full score-path suppression.

\section{Extended Robustness Experiments}
\label{sec:extended_robustness}

\begin{figure}[h]
\centering
\includegraphics[width=0.9\textwidth]{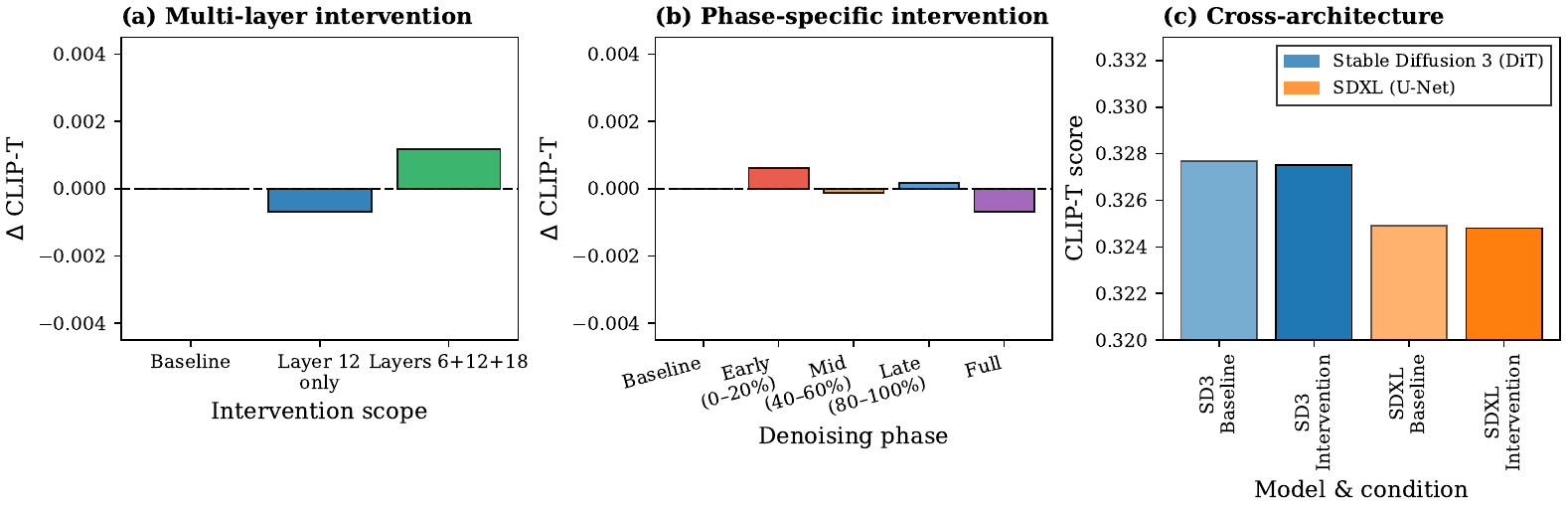}
\caption{\textbf{Robustness analysis.} 
(a) Multi-layer intervention (L6+12+18). 
(b) Phase-specific intervention. 
(c) Cross-architecture validation (SD3 vs SDXL). 
All conditions show no significant quality degradation.}
\label{fig:robustness_appendix}
\end{figure}

This section provides full experimental details for the robustness analyses
summarized in Section~\ref{sec:robustness}; Figure~\ref{fig:robustness_appendix} visualizes the multi-layer, phase-specific, and cross-architecture results.

\subsection{Multi-Layer Intervention}

To test whether removing sinks from a single layer may be compensated by
other layers, we simultaneously intervene on layers 6, 12, and 18 (Table~\ref{tab:multilayer_appendix}).

\begin{table}[h]
\centering
\caption{\textbf{Multi-layer intervention results.} 
Simultaneous removal across three layers produces no significant degradation.}
\label{tab:multilayer_appendix}
\begin{tabular}{lccc}
\toprule
Condition & CLIP-T & $\Delta$ ($\times 10^{-3}$) & $p$ \\
\midrule
Baseline & 0.3288 & --- & -- \\
Single-L (L12) & 0.3281 & $-0.7$ & 0.72 \\
Multi-L (6+12+18) & 0.3300 & $+1.2$ & 0.68 \\
\bottomrule
\end{tabular}
\end{table}

\subsection{Phase-Specific Intervention}

Given that sinks are strongest during early denoising (Section~\ref{sec:h1}), 
one might hypothesize that they serve a critical function specifically in the 
high-noise regime. We test this by applying interventions only during specific
denoising phases (Table~\ref{tab:phase_appendix}).

\begin{table}[h]
\centering
\caption{\textbf{Phase-specific intervention results.} 
Sink removal during any phase, including early denoising where sinks are 
strongest, does not degrade quality.}
\label{tab:phase_appendix}
\begin{tabular}{lccc}
\toprule
Condition & CLIP-T & $\Delta$ & Phase ($t/T$) \\
\midrule
Baseline & 0.3288 & -- & -- \\
Early-only & 0.3294 & $+0.0006$ & $[0.0, 0.2]$ \\
Mid-only & 0.3286 & $-0.0001$ & $[0.4, 0.6]$ \\
Late-only & 0.3290 & $+0.0002$ & $[0.8, 1.0]$ \\
Full removal & 0.3281 & $-0.0007$ & $[0.0, 1.0]$ \\
\bottomrule
\end{tabular}
\end{table}

\section{Robustness and Attribution Analyses}
\label{sec:attribution_appendix}

This section provides full experimental details for the robustness and 
attribution analyses summarized in Section~\ref{sec:robustness_attribution}.

\subsection{E1: Task-Type Robustness}

Using 553 GenEval prompts with non-exclusive semantic tags (single-object,
multi-object, counting, color, position), we compute paired differences
relative to baseline; per-tag CLIP-T shifts and LPIPS are reported in Table~\ref{tab:e1_appendix}.

\begin{table}[h]
\centering
\caption{\textbf{Task-type robustness (E1).} $\Delta_{C}$: Change in CLIP-T 
scaled by $10^3$. Most task types show CIs covering zero; counting exhibits 
a small positive boundary effect.}
\label{tab:e1_appendix}
\vskip 0.1in
\begin{small}
\begin{tabular}{lcc}
\toprule
Task Type & $\Delta_{C}$ ($10^{-3}$) [95\% CI] & LPIPS \\
\midrule
Single-object & $\approx 0$ (CI $\ni 0$) & 0.18 \\
Multi-object & $\approx 0$ (CI $\ni 0$) & 0.19 \\
Counting & $+2.2$ [$+1.0, +4.0$] & 0.18 \\
Color & $\approx 0$ (CI $\ni 0$) & 0.20 \\
Position & $\approx 0$ (CI $\ni 0$) & 0.21 \\
\bottomrule
\end{tabular}
\end{small}
\end{table}

For most tags, 95\% bootstrap CIs of $\Delta$CLIP-T include zero, indicating 
no selective regressions. The counting tag shows a small positive shift 
($\Delta$CLIP-T $\approx +0.0022$), representing a boundary effect rather 
than a regression.

\subsection{E2: Sampling Sensitivity}

We vary classifier-free guidance (CFG), denoising steps, and schedulers
(default, Euler, DPM++) on a fixed prompt subset (Table~\ref{tab:e2_appendix}).

\begin{table}[h]
\centering
\caption{\textbf{Sampling sensitivity (E2).} $\Delta_{C}$: Change in CLIP-T 
scaled by $10^3$. All CIs largely cover zero.}
\label{tab:e2_appendix}
\vskip 0.1in
\begin{small}
\begin{tabular}{lccc}
\toprule
Setting & $\Delta_{C}$ ($10^{-3}$) [95\% CI] & LPIPS \\
\midrule
CFG 3.0 & $\approx 0$ (CI $\ni 0$) & 0.07 \\
CFG 7.5 (default) & $\approx 0$ (CI $\ni 0$) & 0.19 \\
CFG 12.0 & $\approx 0$ (CI $\ni 0$) & 0.31 \\
\midrule
Steps 8 & $\approx 0$ (CI $\ni 0$) & 0.15 \\
Steps 20 (default) & $\approx 0$ (CI $\ni 0$) & 0.19 \\
Steps 50 & $\approx 0$ (CI $\ni 0$) & 0.22 \\
\midrule
Scheduler: Default & $\approx 0$ (CI $\ni 0$) & 0.19 \\
Scheduler: Euler & $\approx 0$ (CI $\ni 0$) & 0.18 \\
Scheduler: DPM++ & $+1.8$ (boundary) & 0.20 \\
\bottomrule
\end{tabular}
\end{small}
\end{table}

Across all tested settings, $\Delta$CLIP-T remains near zero with CIs 
largely covering zero. LPIPS increases with higher CFG, consistent with 
known guidance effects, without affecting alignment.

\subsection{E3: Text vs.\ Image Sink Attribution}

Under SD3's joint-attention formulation, we attribute sinks to token 
categories. Restricting to genuine joint-attention instances
($n_{\text{text}}>0$), sinks occur almost exclusively on text key
positions (47,999/48,000 records; Table~\ref{tab:e3_appendix}).

\begin{table}[h]
\centering
\caption{\textbf{Text vs.\ image sink attribution (E3).} Selective ablation 
confirms that perceptual effects are dominated by text-side sinks.}
\label{tab:e3_appendix}
\vskip 0.1in
\begin{small}
\begin{tabular}{lcc}
\toprule
Ablation Mode & $\Delta_{C}$ ($10^{-3}$) [95\% CI] & LPIPS \\
\midrule
Text-only sinks & $-1.2$ [$-4.0, +1.0$] & 0.160 \\
Image-only sinks & $-0.1$ [$-1.0, +1.0$] & 0.037 \\
All sinks & $-1.2$ [$-4.0, +1.0$] & 0.160 \\
\bottomrule
\end{tabular}
\end{small}
\end{table}

Masking text sinks yields substantially larger appearance changes 
(LPIPS $\approx 0.160$) than masking image sinks (LPIPS $\approx 0.037$), 
while semantic alignment remains preserved (paired $\Delta$CLIP-T CIs 
include zero). The near-exclusive concentration of sinks on text keys 
suggests a recurrent empirical pattern in text–image coupling.

\section{Sanity Checks}
\label{sec:sanity}

\subsection{No-op Implementation Verification}

To ensure that observed perceptual and distributional shifts arise 
from the intervention itself rather than implementation artifacts, 
we conduct two no-op sanity checks.

We generate 100 images under three conditions using identical prompts
and seeds:
\begin{enumerate}[nosep]
    \item \textbf{Baseline}: Original SD3 pipeline, no modifications.
    \item \textbf{Noop wrapper}: Custom attention wrapper installed, 
          but directly calls original processor.
    \item \textbf{Noop processor}: Full \texttt{DynamicSinkProcessor} 
          installed with \texttt{intervention\_enabled=False}.
\end{enumerate}

Both no-op conditions produce outputs that are \emph{pixel-identical}
to baseline (Table~\ref{tab:sanity_check}).

\begin{table}[h]
\centering
\caption{\textbf{No-op sanity check results.} Both no-op conditions produce 
pixel-identical outputs to baseline, confirming that observed shifts arise 
solely from active intervention.}
\label{tab:sanity_check}
\vskip 0.1in
\begin{small}
\begin{tabular}{lccc}
\toprule
Comparison & Pixel Diff & LPIPS & FID$_{\text{shift}}$ \\
\midrule
Baseline vs Noop Wrapper & 0.000 & 0.000 & 0 \\
Baseline vs Noop Processor & 0.000 & 0.000 & 0 \\
\midrule
Baseline vs \textbf{Active Intervention} & 14.5 & 0.18 & 432--996 \\
\bottomrule
\end{tabular}
\end{small}
\end{table}

These results confirm that:
\begin{itemize}[nosep]
    \item Installing custom processors does not introduce artifacts.
    \item The processor correctly passes through when disabled.
    \item All observed shifts (LPIPS $\approx 0.18$, FID $\approx 400$--$1000$) 
          arise \emph{solely} from active sink removal.
\end{itemize}

This verification is critical for establishing causal attribution: 
the perceptual changes reported in Section~\ref{sec:perceptual} are 
caused by the intervention, not by implementation side effects.

To contextualize the observed FID$_{\text{shift}}$ values ($\approx$400--1000),
Appendix~\ref{sec:fid_calibration} reports calibration baselines under 
pure seed variation and common hyperparameter changes.
Notably, seed variation alone (same settings, different random seeds) 
yields FID $\approx 115$, establishing that FID is highly sensitive to 
sampling stochasticity even without any intervention.

\section{FID Calibration Baselines}
\label{sec:fid_calibration}

To interpret the FID$_{\text{shift}}$ values reported in the main text,
we measure FID between image sets generated under common variations
that do not involve any attention intervention (Table~\ref{tab:fid_calibration}).

\begin{table}[h]
\centering
\caption{\textbf{FID calibration baselines (SD3, $N=100$).}
FID shifts from common variations provide reference points for 
interpreting intervention effects.
Seed variation alone produces FID $\approx 115$, establishing the 
stochastic baseline.}
\label{tab:fid_calibration}
\vskip 0.1in
\begin{small}
\begin{tabular}{lc}
\toprule
Comparison & FID \\
\midrule
Seed variation (same settings) & 115.1 \\
CFG 7.5 $\rightarrow$ 6.5 ($\Delta$ = $-1$) & 53.6 \\
CFG 7.5 $\rightarrow$ 8.5 ($\Delta$ = $+1$) & 57.6 \\
Steps 20 $\rightarrow$ 15 ($\Delta$ = $-5$) & 81.3 \\
Steps 20 $\rightarrow$ 10 ($\Delta$ = $-10$) & 108.5 \\
Scheduler: Flow $\rightarrow$ Euler & 330.5 \\
\bottomrule
\end{tabular}
\end{small}
\end{table}

These results show that FID is highly sensitive to sampling stochasticity: 
generating images with identical settings but different random seeds 
produces FID $\approx 115$.
Common hyperparameter variations (CFG $\pm 1$, fewer steps) yield FID 
in the range of 50--110, while scheduler changes can produce much larger 
shifts (FID $\approx 330$).

The intervention-induced FID$_{\text{shift}}$ values reported in the 
main text ($\approx 400$--$1000$) are thus comparable in magnitude to 
aggressive hyperparameter changes or scheduler substitutions, rather 
than indicating anomalous distributional collapse.
This calibration supports the interpretation that sink suppression 
moves samples within the model's output manifold without fundamentally 
altering the generation process

\section{Sink-Specificity Analysis Under Stronger Interventions}
\label{sec:sink_specificity}

\subsection{CLIP-T Budget Sweep}

To verify alignment robustness across masking intensities, we evaluate 
CLIP-T under counterfactual ablation at $k \in \{1, 5, 10, 20, 50\}$ 
(layer 12, $N{=}64$).

\begin{table}[h]
\centering
\caption{\textbf{CLIP-T under varying masking budgets.} 
All conditions show CI$\ni$0 after Holm correction.}
\label{tab:clip_k_sweep}
\vskip 0.1in
\begin{small}
\begin{tabular}{lrrrl}
\toprule
$k$ & Mode & $\Delta$ & 95\% CI & $p_{\text{adj}}$ \\
\midrule
1 & top\_sink & $-0.001$ & $[-0.003, +0.002]$ & 1.00 \\
5 & top\_sink & $-0.001$ & $[-0.004, +0.002]$ & 1.00 \\
10 & top\_sink & $-0.001$ & $[-0.005, +0.003]$ & 1.00 \\
20 & top\_sink & $-0.001$ & $[-0.006, +0.003]$ & 1.00 \\
50 & top\_sink & $-0.002$ & $[-0.008, +0.004]$ & 1.00 \\
\bottomrule
\end{tabular}
\end{small}
\end{table}

Across all budgets, CLIP-T changes remain within the practical equivalence 
range with 95\% CIs including zero, confirming that alignment is robust 
to sink removal regardless of intervention intensity.

\subsection{Sink-Specificity Test (HPS-v2)}

To assess whether preference effects are \emph{sink-specific}, we compare 
sink masking against equal-budget random masking using a paired 
difference-of-differences test:
\[
d_i = (\text{HPS}_{\text{sink},i} - \text{HPS}_{\text{base},i}) - 
      (\text{HPS}_{\text{rand},i} - \text{HPS}_{\text{base},i})
\]
We test $\Delta\Delta = \mathbb{E}[d_i] < 0$ (one-sided).

\begin{table}[h]
\centering
\caption{\textbf{Sink-specificity under HPS-v2} (layer 12, $N{=}64$). 
$\Delta\Delta = \Delta_{\text{sink}} - \Delta_{\text{rand}}$ (one-sided test, $\Delta\Delta < 0$).
At $k{=}1$, no sink-specific effect is observed. At $k \geq 10$, 
sink masking degrades HPS-v2 significantly more than random masking.
Significance is determined by 95\% bootstrap CI excluding zero.}
\label{tab:sink_specificity_appendix}
\vskip 0.1in
\begin{small}
\begin{tabular}{lrrrr}
\toprule
$k$ & $\Delta\Delta$ & 95\% CI & $p$ (one-sided) & \\
\midrule
1 & $-0.002$ & $[-0.005, +0.002]$ & 0.16 & n.s. \\
10 & $-0.005$ & $[-0.008, -0.001]$ & 0.007 & sink-specific \\
50 & $-0.020$ & $[-0.026, -0.013]$ & $<10^{-4}$ & sink-specific \\
\midrule
\multicolumn{5}{l}{\textit{Trend: $\Delta d = d_i(k{=}50) - d_i(k{=}10)$}} \\
\midrule
$50-10$ & $-0.015$ & $[-0.022, -0.008]$ & $<10^{-4}$ & dose-dependent \\
\bottomrule
\end{tabular}
\end{small}
\end{table}

\subsection{Interpretation}

These results reveal a dose-dependent transition in sink-specificity:
\begin{itemize}[nosep]
    \item \textbf{$k{=}1$ (standard setting)}: No sink-specific effect. 
          Both sink and random single-token removal are tolerated 
          ($\Delta\Delta$ CI includes zero). This confirms that 
          non-necessity is not an artifact of ``any token'' being removable.
    \item \textbf{$k \geq 10$ (stronger intervention)}: Sink-specific 
          degradation emerges in HPS-v2, with effect magnitude increasing 
          across tested budgets ($k{=}10, 50$).
    \item \textbf{CLIP-T}: No sink-specific effect at any tested budget. 
          Alignment remains robust regardless of intervention intensity.
\end{itemize}

Our primary non-necessity claim (Section~\ref{sec:dynamic_sink}) concerns 
standard inference settings ($k{=}1$); these results delineate boundary 
conditions under stronger-than-mainline interventions.

\subsection{Qualitative Comparison Panels}
\label{sec:qualitative_panels}

\begin{figure*}[t]
\centering
\includegraphics[width=0.9\textwidth]{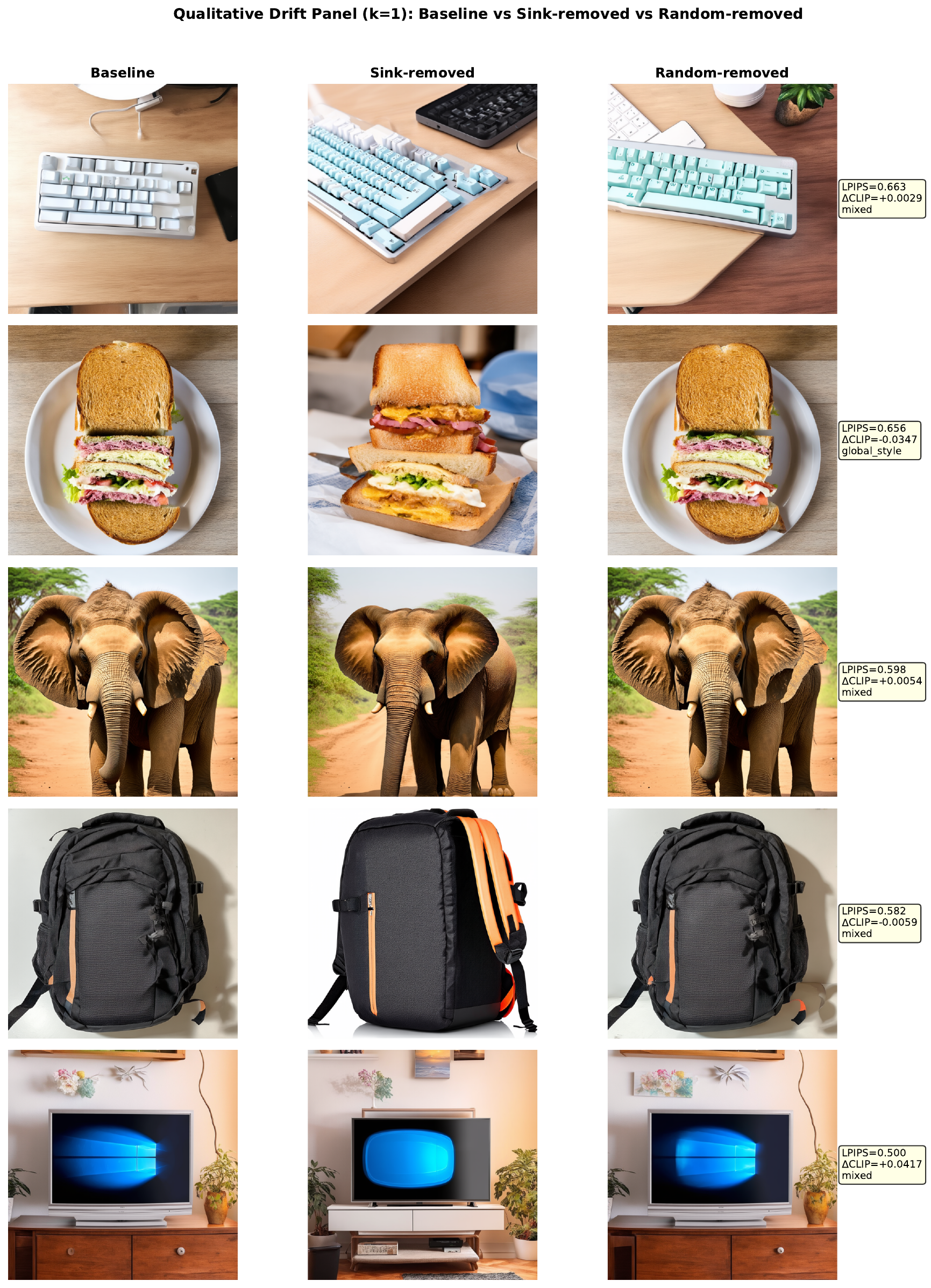}
\caption{\textbf{Qualitative drift comparison at $k{=}1$ 
(union-budget protocol, layer~12).} Each row shows the same prompt and 
seed under three conditions: baseline (left), sink-removed (middle), 
and equal-budget random-removed (right). Per-image annotations report 
LPIPS and $\Delta$CLIP-T relative to baseline. Sink masking 
consistently produces larger appearance changes (layout, color, viewpoint) 
than random masking while preserving the prompted concept.}
\label{fig:qualitative_k1}
\end{figure*}

\begin{figure*}[t]
\centering
\includegraphics[width=0.9\textwidth]{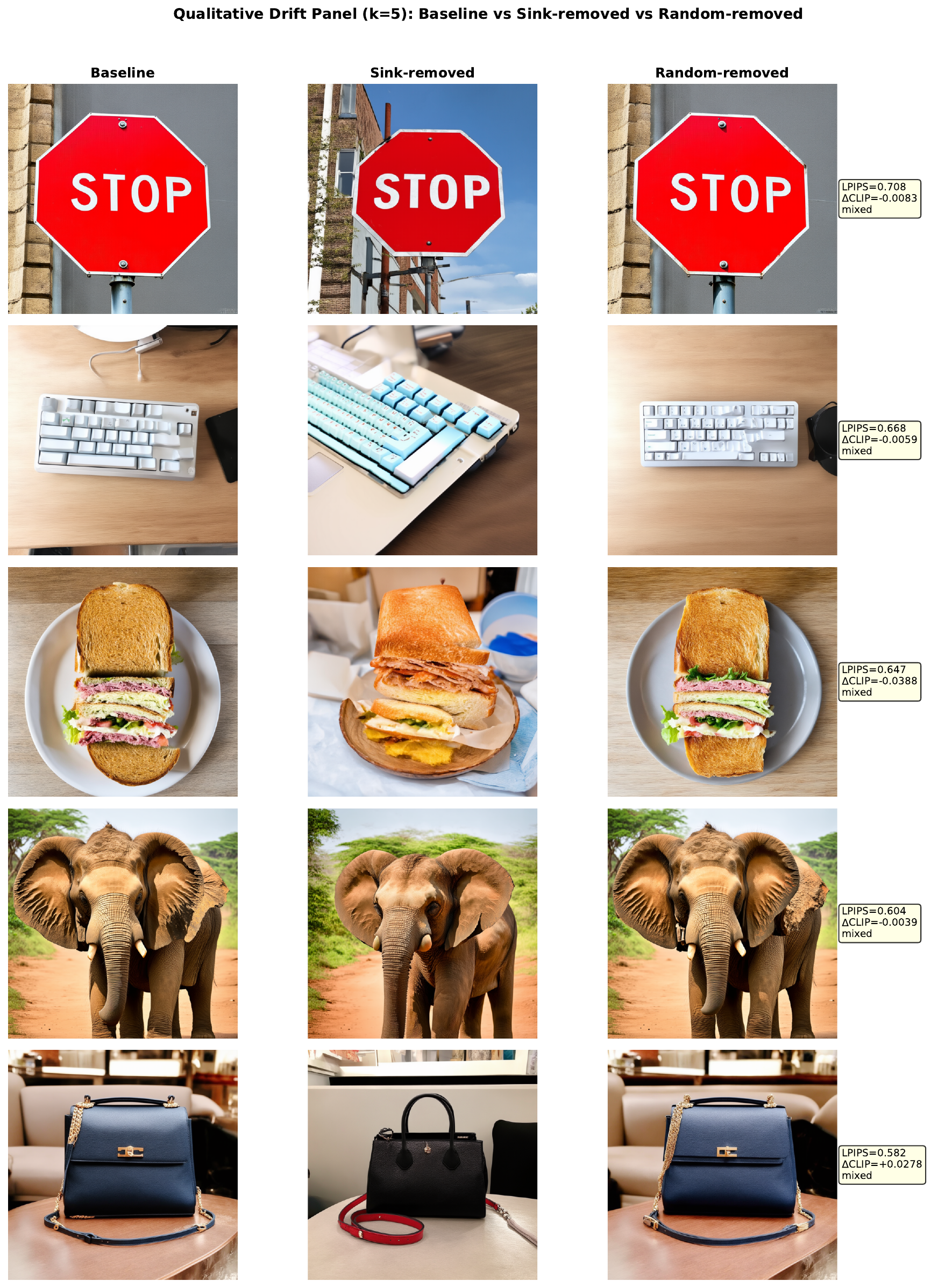}
\caption{\textbf{Qualitative drift comparison at $k{=}5$.} 
Same layout as Figure~\ref{fig:qualitative_k1} with a stronger 
intervention budget. The sink-vs-random visual gap widens, consistent 
with the larger $\Delta\Delta$ LPIPS at $k{=}5$ in 
Table~\ref{tab:sink_vs_random_lpips}.}
\label{fig:qualitative_k5}
\end{figure*}

To visualize the dissociation documented quantitatively in 
Table~\ref{tab:sink_vs_random_lpips}, Figures~\ref{fig:qualitative_k1} 
and~\ref{fig:qualitative_k5} show side-by-side comparisons 
(baseline vs.\ sink-removed vs.\ random-removed) for representative 
prompts at $k{=}1$ and $k{=}5$. Sink-removed outputs exhibit substantial 
layout, viewpoint, and style restructuring while preserving the prompted 
coarse concept; random-removed outputs remain close to baseline. The 
visual gap between the two ablation conditions is consistent with the 
$\sim\!6\times$ LPIPS difference reported in 
Section~\ref{sec:sink_specificity_main}.

\end{document}